\documentclass{article}

\usepackage{arxiv}

\usepackage[utf8]{inputenc} 
\usepackage[T1]{fontenc}    
\usepackage{hyperref}       
\usepackage{url}            
\usepackage{booktabs}       
\usepackage{amsfonts}       
\usepackage{nicefrac}       
\usepackage{microtype}      
\usepackage{lipsum}
\usepackage{graphicx}
\graphicspath{ {./images/} }

\title{Modelling the COVID-19 virus evolution with Incremental Machine Learning}

\author{
 Andrés L. Suárez-Cetrulo \\
  Ireland's Centre for Applied AI (CeADAR).\\
  University College Dublin. Ireland\\
  \texttt{andres.suarez-cetrulo@ucd.ie} \\
   \And
 Ankit Kumar \\
  Ireland's Centre for Applied AI (CeADAR).\\
  University College Dublin. Ireland\\
  \texttt{ankit.kumar@ucd.ie} \\
  \And
 Luis Miralles-Pechuán \\
  School of Coumputing\\
  Technological University Dublin. Ireland\\
  \texttt{luis.miralles@tudublin.ie} \\
}

\begin{document}
\maketitle
\begin{abstract}
The investment of time and resources for better strategies and methodologies to tackle a potential pandemic is key to deal with potential outbreaks of new variants or other viruses in the future. In this work, we recreated the scene of a year ago, 2020, when the pandemic erupted across the world for the fifty countries with more COVID-19 cases reported. We performed some experiments in which we compare state-of-the-art machine learning algorithms, such as LSTM, against online incremental machine learning algorithms to adapt them to the daily changes in the spread of the disease and predict future COVID-19 cases.

To compare the methods, we performed three experiments: In the first one, we trained the models using only data from the country we predicted. In the second one, we use data from all fifty countries to train and predict each of them. In the first and second experiment, we used a static hold-out approach for all methods. In the third experiment, we trained the incremental methods sequentially, using a prequential evaluation. This scheme is not suitable for most state-of-the-art machine learning algorithms because they need to be retrained from scratch for every batch of predictions, causing a computational burden.

Results show that incremental methods are a promising approach to adapt to changes of the disease over time; they are always up to date with the last state of the data distribution, and they have a significantly lower computational cost than other techniques such as LSTMs.

\end{abstract}


\section{Introduction}

The coronavirus disease 2019, better known as COVID-19, became knowledge of the public domain in December 2019. Its origin has been placed in Wuhan, in the Hubei province of China \cite{shereen2020covid}. Shortly after its announcement, on January 30 2020, the World Health Organization (WHO) declared the virus a Public Health Emergency of International Concern \cite{zu2020coronavirus}. The COVID-19 is a new virus with a high transmissibility level that causes the severe acute respiratory syndrome coronavirus 2 (SARS-CoV-2). The COVID-19 virus is likely to have its origin in bats since its structure is similar to other well-known coronavirus presented in these animals. It is expected that a human initially contracted the virus through the consumption of one of these infected animals \cite{shereen2020covid}.

Due to its pathogenicity --the capacity of causing harm-- and its ease of transmission across humans, the COVID-19 virus has caused tremendous damage to public health, initially in China and subsequently in the rest of the world. On April 1, 2021, the official numbers of deaths were almost 3 million (2,830,488) \cite{Covid-pandemic}. On top of that, the restrictive actions such as lock-downs taken by governments to mitigate the virus had catastrophic consequences on the economy \cite{bonaccorsi2020economic}. Millions of jobs have been lost, thousands of companies have closed, and many institutions have gone bankrupt \cite{nicola2020socio}.

In April of 2021, seven vaccines were approved and administered massively in many countries such as Israel, U.S., Bhutan, U.K., or Bahrain. At that point, those countries administered more than 40 vaccination doses per 100 people ~\cite{Vaccinated-rate}. However, even if we overcome the COVID-19 pandemic in the next few months, other pandemics can emerge and affect the future population. That is why it is vital to find solutions for addressing a pandemic before it happens again \cite{shereen2020covid}.

Although the virus is mainly addressed from a medical point of view, there are a lot of other disciplines that can positively contribute to mitigating its effects \cite{lalmuanawma2020applications}. In this regard, we are particularly interested in developing Artificial Intelligence (AI) applications to combat the virus. AI has been used in this regard in many ways, such as fast diagnosis and screening processes, contact tracing, vaccine development, and prediction and forecasting effects \cite{lalmuanawma2020applications,wynants2020prediction}.

Our contribution falls in the category of predicting and forecasting the detected positive number of COVID-19 cases. Accurately predicting the number of cases a few weeks in advance allows governments to plan, develop better strategies, and take preventive actions before it is too late \cite{miralles2020methodology}. Imposing restrictive measures may provoke catastrophic effects on the economy while not controlling the virus can cause a high number of deaths \cite{miralles2020methodology}. Therefore, it is essential to develop adequate tools to help governments take the best possible actions to address potential pandemics.

Thanks to predictive models able to make accurate estimations, resources in hospitals can be managed more intelligently, saving more lives; universities can develop strategies for the students' academic year; and drastic measures for the population such as lock-down or self-isolation, with multiple side-effects, could be avoided \cite{nicola2020socio}.

There are two main categories for estimating the evolution of the virus: the compartmental models and the time series (or curve-fitting) models \cite{harvey2020time}. The compartmental models represent the virus evolution using mathematical models that divide the population into different groups \cite{brauer2008compartmental}. Those models can be parameterised to consider various scenarios, but they are susceptible to minor changes and difficulty in understanding. They also require a deep study of the variables, and figuring out the relationship among them implies a long time. They include an important subcategory called Agent-based models that create virtual simulations of the population based on emulating real individuals behaviour by small independent classes called agents \cite{chang2020modelling}.

The other category of models is called time-series models \cite{maleki2020time}. They use historical data of the virus's evolution, such as the registered positive cases per day. These records implicitly consider a high number of variables that are very difficult to understand and measure even for an expert. Some of those factors are: the mental health of the population; the influence of social networks, the variability in the temperatures; the number of lighting hours; the accumulative fatigue of a lock-down; the higher or lower transmitability rate of a new strain, or the unemployment rate of the country. These factors obviously have a relationship to the spread of the virus but it is very difficult to calculate how much they impact on the spread of the virus. The time series of the reported cases is nothing but the combination of all those factors. Time-series models have been proved to work well, and they have an overall advantage over compartmental models; their simplicity in being modelled \cite{harvey2020time}.

Another way to handle these forecasts is with the use of Machine Learning algorithms. Many of these methods can capture nonlinear relationships in the input data with no prior knowledge. For instance, ensembles like Random Forest have obtained state-of-the-art results in complex domains like this \cite{zoabi2021machine, BALLINGS20157046}. Machine Learning (ML) algorithms are starting to replace time-series models as the de-facto standard to make predictions in the literature \cite{Hsu2016}. 
One of the inconveniences of state-of-the-art ML methods is their inability to deal with data updates and their evolution efficiently, in either stationary or non-stationary domains where a hidden context may influence the predictive model behaviour over time in unforeseen ways~\cite{Suarez-Cetrulo2017}. Changes in this hidden context are what the literature of online machine learning algorithms (a sub-field of ML) knows as concept drift~\cite{bahri2021, Gama2014}.  From here on-wards in this paper, we refer to state-of-the-art machine learning algorithms as static algorithms.  

Pandemic curves are non-stationary by nature. Depending on the period, they can show clear trends, cycles, seasons where the random component is more prevalent, etc. Moreover, the virus spread in each region is affected by external factors not captured in the available data \cite{wang2020inference, 10.1093/cid/ciaa418}. Under these circumstances, incremental and online machine learning techniques \cite{Ditzler2015} that adapt to the evolution of the trend and its changes, 
are gaining traction in different domains \cite{Gama2014, e21010025}.
To our knowledge, these methods and the notion of concept drift have not gained enough attention yet in the coronavirus prediction domain. However, incremental machine learning algorithms can deal actively or passively~\cite{Elwell2011IncrementalEnvironments.} 
with the non-stationary nature of data streams such as the COVID-19 curve evolution~\cite{Tsymbal2004TheWork}. This is by adapting (passive) to the non-stationary nature of the data or through the use of drift detectors (active). These methods can find an equilibrium between prioritising new knowledge, adapting to changes, and retaining previous relevant information through different forgetting mechanisms. This balance is known as the stability-plasticity dilemma and is very suitable for dynamics scenarios such as COVID-19 spreads~\cite{incrementalLearningHealth, 10.1007/978-3-540-39907-0_49}. 

This work aims to forecast the number of positive COVID-19 cases in multiple countries using incremental machine learning methods and compare their performance with state-of-the-art methods.
Our contribution consists of proposing a framework to predict the number of new cases while dealing with the evolution in the spread of the curve in different countries. We created this framework (see the link to Github on the first page) to encourage the scientific community to 
develop new models and new strategies to design more accurate algorithms to predict COVID-19 cases.
To our knowledge, no other publications are showing the benefits of applying incremental methods to predict the number of cases in a pandemic, and that makes of our piece of research a significant contribution to this area.

The rest of the paper is organised as follows: In section \ref{state-of-the-art} we present an overview of the models to represent the COVID-19 pandemic and other viruses, as well as an overview of the supervised machine learning models. Particularly a subset of them called incremental methods. In section \ref{experiments}, we present the conducted experiments to compare the performance of the incremental methods against other popular methods such as Long short-term memory networks. Then, we compare the performance of the methods under different scenarios and training schemes to find out the optimal configuration. In section \ref{Discussion}, we compare the obtained results for each of the models and present an analysis for both the static and the incremental methods. Finally, in section \ref{conclusion} we present the main findings of our investigation and recommend some interesting lines of research for future work.

\section{State of the art}\label{state-of-the-art}
This section describes some of the most important research works related to modelling the spread of the COVID-19 Virus on a population. We also include a description of the main ML methods used, emphasising online incremental machine learning algorithms and differences in model evaluation with state-of-the-art static methods.

\subsection{Modelling the evolution of COVID-19}
The impact of the pandemic on society has pushed researchers to find ways to combat the pandemic from their respective disciplines. The volume of publications related to COVID-19 in the last two years is monumental. To give an example, Wynants, Laure et al. \cite{wynants2020prediction} published a study in which they screened 14,217 titles and analysed 107 studies related to models for predicting diagnosis and prognosis of COVID-19; and that is just one of many areas.

Our investigation falls into the category of modelling the evolution of the virus in the population. On this topic, there is a high number of publications covering a broad spectrum that goes from those applying the simple method ARIMA to those implementing the latest deep learning techniques \cite{zeroual2020deep}. Forecasting with precision the evolution of the positive cases or the utilization of beds at hospitals is at the backbone for planning the optimal governmental actions to avoid damage to the economy and protect public health \cite{zeroual2020deep,covid2020optimization}. Some investigations point out how some of the models could mislead governments for not taking the optimal actions and stress the importance of generating reliable predictive models \cite{chin2020case}.

The importance of accurately predicting the spread of the COVID-19 virus is such that many universities created very strong research groups focused on modelling the virus. For instance, the University of Texas created a group called UT Austin COVID-19 Modeling Consortium \cite{AustinCovid-19} that comprises experts from many disciplines such as physics, medicine, biology, maths, computer science and so on, and counts with a large number of resources and publications.

The models that estimate the behaviour of the virus in the population can be grouped, broadly speaking, into two categories; the compartmental and the time-series models. The compartmental models divide the population into dynamic groups using differential equations. The time-series models, also called curve-fitting models, are based on sequential historical data points to predict future records. Compartmental models typically use time-series models to adjust the parameters of the equations. 
 
A common approach in the compartmental models is modelling the virus using a SIR model that divides the population into three groups: Susceptible (S), Infected (I), and Recovered (R). Or an improved version of the SIR model called SEIR that includes a new group called Exposed(E). Then by using just three parameters: $\beta$ (contact rate), $\sigma$ (incubation period) and $\gamma$ (recovery rate or period in which individuals are infectious) \cite{yang2020modified}. Differential equations model the transition of individuals from one to other group. There is even an extension of the SEIR model called SEIRS that includes a new parameter representing the individuals in the Recovered group that moved to the Susceptible group by losing the immunity after a given time. This factor is modelled by a variable called $\epsilon$ \cite{yang2020modified}.

\begin{figure}
  \centering
  \includegraphics[width=0.7\linewidth]{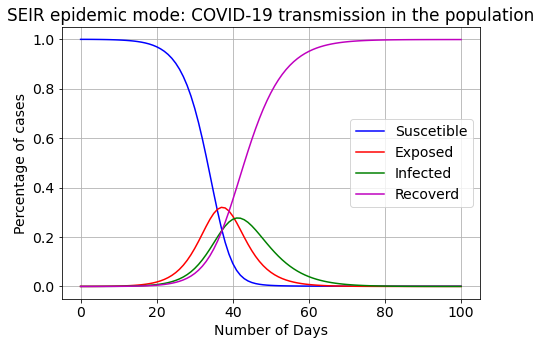}
  \caption{Simulation of the spread of the COVID-19 virus in the population with the following parameters: $\beta=1.1$, $\alpha=0.19$, $\gamma = 0.18$ and 0.03\% infected population.}
  \label{fig:evaluation}
\end{figure}

Compartmental models define the behaviour of the groups (S,E,I,R) by differential equations that consider many factors \cite{brauer2008compartmental}. Some of those factors are the population's size and age, whether there is a lockdown, whether people respect the two-meter distance or use masks. There are also other factors related to the virus such as the level of propagation, the incubation period and so on. Compartmental models have the caveat that a large number of assumptions represent a changing environment where small details can render the model flawed. The virus can mutate, becoming more harmful or more contagious \cite{toyoshima2020sars} which makes necessary continuously adjusting the model. On top of that, the more complex the models are, the more difficult to understand. The main advantage is that it enables creating simulations for different scenarios.

There is an important subcategory inside the compartmental models called Agent-based models. Agent-based models approximate the real-world scenario by representing the population's individuals using virtual agents inside of a software platform. Concerning this subcategory, a relevant approach represents the whole population of a country with virtual individuals. For example, at the University of Australia, Chang et al. \cite{chang2020modelling} implemented a fine-grained simulation and calibrated the model to match the COVID-19 transmission rate in Australia. They studied the impact of some factors such as air travel, case isolation, or home quarantine. And they concluded that closing schools is not a decisive factor to combat the virus. Using these methods, some researchers created very complex algorithms to describe the evolution of the COVID-19 and its impact on the population's economy and public health. Another example of this category is that of Silva, Petrônio CL, et al. \cite{silva2020covid}. In their investigation they propose a method based on agents called COVID-ABS (Agent-based system). Their method simulates the behaviour of the pandemic using agents to represent people, business and governments. They also consider different scenarios such as lockdown, using face masks, partial isolation, doing nothing, or partial isolation. They tried to find a balance between damaging the economy and avoiding a high number of deaths.

The approach we are considering in our research was also used by the Institute for Health Metrics and Evaluation (IHME) at the University of Washington; this is based on curve-fitting. The institute predicted quite accurately when the United States had a peak on the curve of the number of cases by creating models that fitted the curve of positive cases of the virus in other countries such as Italy, Spain, or the UK \cite{covid2020forecasting}. Other research works, such as Zeroual, Abdelhafid et al. \cite{zeroual2020deep} presented a general overview of some of the most important studies on modelling the virus and compared five different deep learning methods very suitable for modelling time-series data such as simple Recurrent Neural Network (RNN), Long short-term memory (LSTM), Bidirectional LSTM (BiLSTM), Gated recurrent units (GRUs) and Variational AutoEncoder (VAE). One of the downsides of these models is that they are not configurable. Therefore, it is not possible to create simulations to consider different scenarios in which you can tune the variables and see how the virus changes.

Online incremental algorithms have not been applied yet to predict the cases of Coronavirus disease. We believe that our work will help to fill that gap in the literature. These methods are widely used in non-stationary domains like financial transactions, telecommunications, weather forecast and the Internet Of Things, to name a few, 
adapting to shifts and drifts that may occur implicitly in the data~\cite{Gama2014}. 
That is why we think they can be a good solution to model the COVID-19 curves.
\subsection{Supervised Machine Learning models and Incremental algorithms}

In the following subsections, we briefly describe the regression algorithms and the evaluation schemes implemented in our experiments.

\subsubsection{Static Machine Learning Algorithms}
We selected some of the most relevant regression machine learning models to compare them with the incremental methods. The description of the algorithms chosen goes as follows.
\begin{itemize}
\item Linear regression (LR) minimises the residual sum of squares between the prediction and the target feature and fits a line with coefficients \cite{montgomery2012introduction}.
\item Ridge regression, as an improvement of LR, uses regularization, minimises $\beta$ coefficients and adds a $\lambda$ scalar to the learning process \cite{bishop2006pattern}. Bayesian Ridge Regression \cite{mackay1992bayesian, tipping2001sparse} is the Bayesian interpretation of a Ridge Regression Estimator. Thus, it performs linear regressions through probability distributors rather than point estimates.
\item Support Vector Machines (SVM) construct a hyperplane through Support Vectors to use it as a discriminator for classification tasks. Support Vector Regression (SVR) is the regression version of the SVM algorithm \cite{chang2011libsvm}.
\item Random Forest (RF) is a popular ensemble method that constructs a set of decision trees through bagging, and feature bagging \cite{liaw2002classification}. The final prediction depends on voting or an aggregation mechanism. The regression version aggregates them by averaging their predictions.
\item Gradient Boosting trains its base learners gradually and sequentially \cite{friedman2001greedy, hastie2009elements}. It uses gradients in the base learners' loss function to measure the outcome of each observation and improves weak learners iteratively.
\item Long Short-Term Memory neural networks (LSTM) \cite{hochreiter1997long} is an architecture with dense and LSTM layers. LSTMs are used to keep adjacent temporal information whilst remembering information for a long time in their memory blocks. In feed-forward neural networks like them, learning occurs by changing these connection weights, often through a gradient descent-based approach like the back-propagation algorithm, to minimise the obtained error \cite{murtagh1991multilayer, glorot2010understanding}. 
\end{itemize}

For these algorithms, we have used their implementations from Scikit-learn, except for the LSTM, where we have used Keras.

\subsubsection{Incremental Machine Learning Algorithms}
We selected a set of four incremental regression machine learning models to compare them with the state-of-the-art methods aforementioned. Our choice covers incremental decision trees that are notoriously used for regression problems in the literature~\cite{bahri2021} and Adaptive Random Forest for Regression~\cite{Gomes2017}, an ensemble of incremental trees, with state-of-the-art results in online incremental learning. A description of these can be seen below.

\begin{itemize}
    \item A Hoeffding tree (HT) is an incremental algorithm that assumes that the data distribution is constant over time. This relies mathematically on Hoeffding bounds, which support the fact that a small sample may suffice to choose an optimal splitting attribute. Hoeffding Trees for regression calculate the decrease of the variance of the target to decide the splits. Its leaf nodes fit linear perceptron models by default~\cite{domingos2000mining}.
    \item The Hoeffding Adaptive Tree (HAT) is an adaptive version of the Hoeffding Tree. It replaces old branches with new ones if the error of the old ones increases over time and new branches perform better. To monitor the evolution of the errors, it uses the Adaptive Windowing (ADWIN) algorithm \cite{bifet2009adaptive, bifet2007learning}. HAT also proposes a bootstrap sampling as an improvement over Hoeffding Trees. 
    \item Adaptive Random Forest (ARF) \cite{Gomes2017} is an Adaptive version of the Random Forest ensemble for Data Streams. 
    It manages a pool of trees that are replaced with new ones when a Concept Drift is detected. As an improvement of RF, each adaptive tree is trained with different samples and feature sets as part of the bagging and the feature bagging process. 
    \item The Passive-Aggressive algorithm (PA) is an online learning algorithm that updates the model depending on the obtained error \cite{JMLR:v7:crammer06a}. 
\end{itemize}

In this paper, we use the version of Passive-Aggressive Regressor provided in Scikit-Learn \cite{pedregosa2011scikit}. For the rest of the approaches, we have used the implementation provided in Scikit-multiflow.

\subsubsection{Differences in model evaluation standards}
One of the main differences when comparing static with incremental machine learning models for data streams is the convention to measure the model performance. Different model evaluation metrics creates a challenge when comparing static with incremental models as the evaluation for both methods needs to be consistent and fair.

Hold-Out is a typical evaluation technique in ML where there are a clear train and test set split. Train and test sets could be split into different fashions, but the main idea is that a hold-out evaluation separates the original set into two independent sets for training and testing. The idea is that the instances from the training set are different from those in the testing set to evaluate how accurate the model can predict. Using 80\% for training and 20\% for testing is a very common split. Hold-out is the evaluation scheme performed in traditional machine learning approaches  \cite{gama2013evaluating}.

A prequential evaluation (or Interleaved Test-Then-Train) is a conventional setting in online (incremental) machine learning for continuous data streams where data is continuously evaluated as soon as it is available \cite{Cerqueira2020}. In this evaluation technique, every data example (instance) is used initially for making a prediction. Then, when its target label is available, the instance is first used to compute the prediction error and, later, to update the algorithm. This differs from the hold-out evaluation, 
where the testing splits are not used for training. 
The prequential evaluation makes more efficient use of the data \cite{Cerqueira2020}, and it is suitable for incremental methods that can be updated and adapted to new instances and, unlike static methods, do not need to retrain the whole model from scratch \cite{Zliobaite2015}.

\section{Experiments and results}\label{experiments}

In this section, we conduct some experiments in which we compare the performance of incremental machine learning methods with that of state-of-the-art static methods; among them, we emphasise the popular deep learning method called LSTM.  Our goal is to see if incremental learning techniques can quickly adapt to the COVID-19 spread for predicting the number of cases in each country.

\subsection{Dataset description}

For this work, we used the dataset "COVID-19 Coronavirus data - daily (up to December 14th 2020)" available in the European Open Data Portal\footnote{https://data.europa.eu/euodp/en/data/dataset/covid-19-coronavirus-data-daily-up-to-14-december-2020} and provided by the European Centre for Disease Prevention and Control. The original dataset contains twelve columns with daily information about the disease in 213 countries during 2020; it is structured as follows. One column represents the number of positive cases; another column the number of deaths, four columns are related to the current date, four other columns to country-specific information, one column refers to the continent, and lastly, there is one column to represents the cumulative number of the COVID-19 cases for 14 days per 100,000 inhabitants.

Regarding the preprocessing steps performed before creating the final dataset for the experiments, the columns related to dates and countries were used to split the training and testing datasets. The number of new cases is the only column used to generate the feature set (input of the model) and the target (output of the model). The rest of the columns were removed.

\begin{figure}[ht]
  \centering
  \includegraphics[width=0.75\linewidth, trim=38 15 13.5 0, clip]{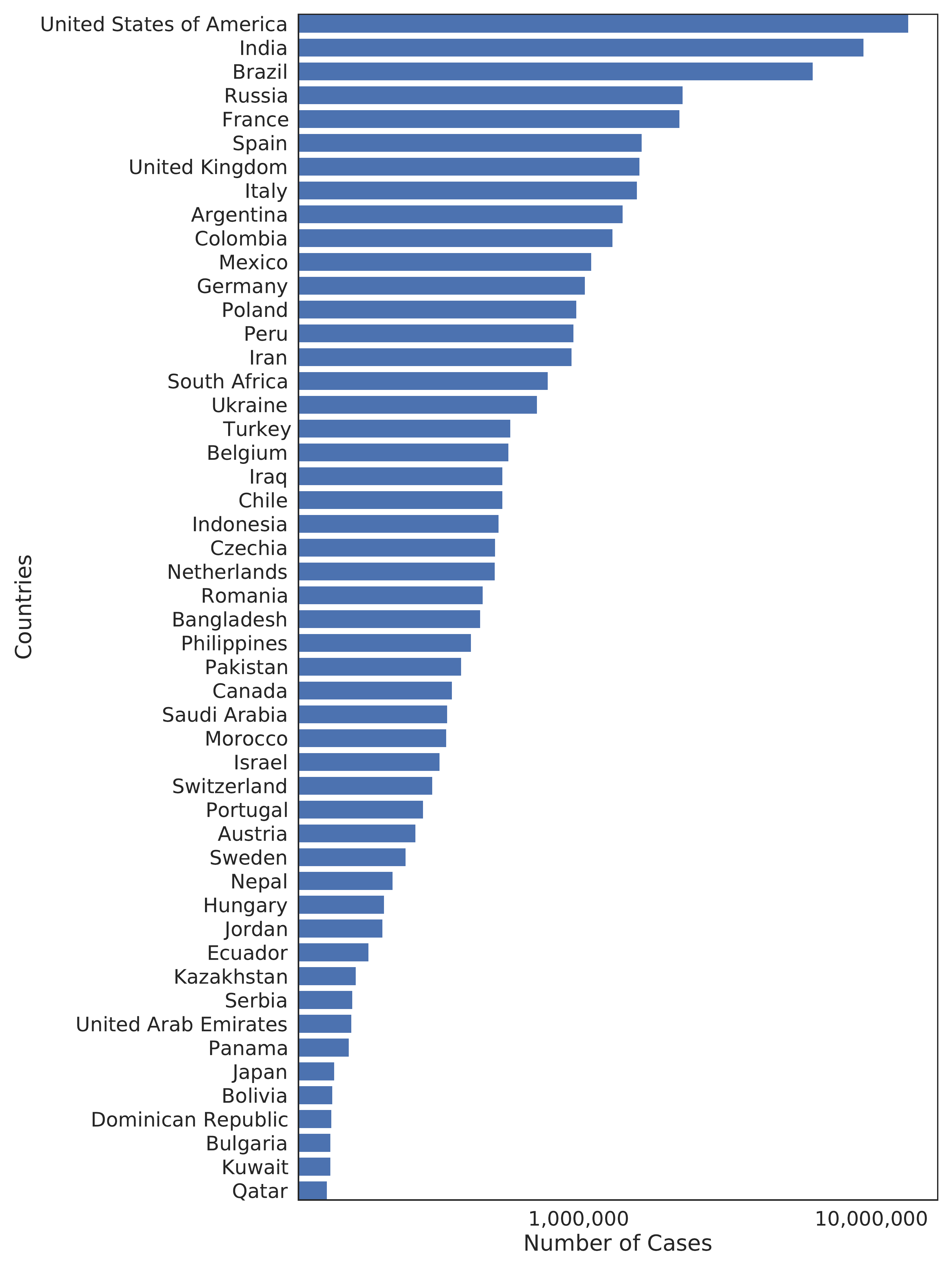}
	\caption{Set of 50 countries with 8 or more months of data and greatest number of cases by the 1st of December of 2020.}
	\label{fig:country_selection}
\end{figure}

Of the countries with eight or more months of data by November 30th, 2020 (66 in total), we selected the 50 with the highest number of accumulated cases of COVID-19 to conduct the experiments. These countries can be seen in Figure \ref{fig:country_selection}.

Each data example corresponds to a moving time window of fifty consecutive days representing the inputs of the regression model, and the target/output of the model is the average of ten consecutive days, where the first of those ten days is 30 days ahead of the last day of the input. The main reason for using the average of ten days is to average some spikes due to potential delays when reporting the test results. We also wanted to predict 30 days ahead because it allows governments to plan the next few weeks in terms of lifting or applying new restrictions on the population.

\begin{figure}
  \centering
  \includegraphics[width=0.85\linewidth]{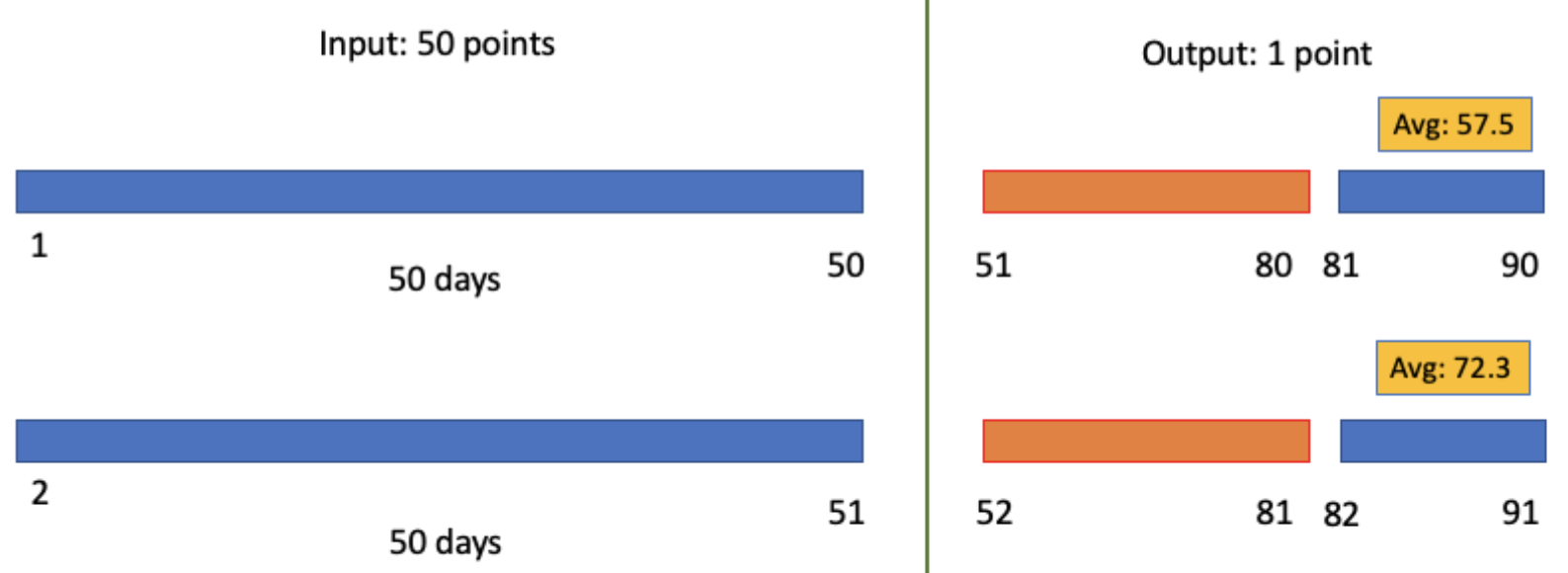}
  \caption{Fifty points are taken as the models’ input to predict a single point, which is the average of ten consecutive days after 30 days.}
  \label{fig:dataset}
\end{figure}

To illustrate how the dataset was transformed, we provide the following example. Imagine we had 200 hundred days of positive COVID-19 new cases for a given country. Then, the first row of the dataset has as the input from day 1 to day 50, and as the output, the average from day 80 to 90; the input for the second row goes from day 2 to 51, and the output the average from 81 to 91; and so on until we there are no more days in the original dataset. As a result, the generated dataset has 50 columns representing the number of new cases in the previous days and a single value representing the output of the model. The number of rows for each trained model varies according to the experiment, as explained in more detail in section \ref{section:experiments}. The feature set (inputs of the model) was created according to the scheme shown in Figure~\ref{fig:dataset}. 

\subsection{Experimentation Methodology}

Our primary concern in these experiments is to put ourselves in the shoes of a country facing a pandemic that only has the information available at a certain date and that needs to generate models to predict future cases of COVID-19 so that it can provide the government with information for taking the optimal actions. Those actions could be closing schools, limiting public transport, applying lock-downs, among others.

The experiments were conducted to answer the following two questions. First, which methods, between the incremental methods and the static ones, have higher performance for predicting the number of new COVID-19 cases? And, second, what is the best option between these two for training a model to predict the future number of cases of COVID-19 for a given country: a) training the model only with the samples of that same country which the model was going to predict, or b) training the model with the complete dataset of fifty countries?

To respond to these questions, initially, we performed three experiments. In experiment I, we trained the static and incremental methods with only one country, and we predicted the future COVID-19 cases for that same country. And in the second experiment, we predicted over one country, but this time, we trained the supervised models with the 50 countries with most cases as shown in Figure \ref{fig:country_selection}. 
Then, we compared the performance of training the models using a single-country with the results obtained using multiple countries.

To make incremental learners comparable, we trained and tested all the algorithms using the same training and test sets. However, incremental learners for data streams are designed to be trained continuously, and static train and test splits are not generally applied to incremental methods in the literature. Thus, to follow the convention with this type of learners, we performed a third experiment to measure the effect of using a unique test and train set only for the incremental learners (prequential evaluation \cite{Cerqueira2020, 10.5555/1855075}), rather than using defined training and test splits.

\begin{figure}[h]
  \centering
  \includegraphics[width=0.85\linewidth, trim=85 540 125 80, clip]{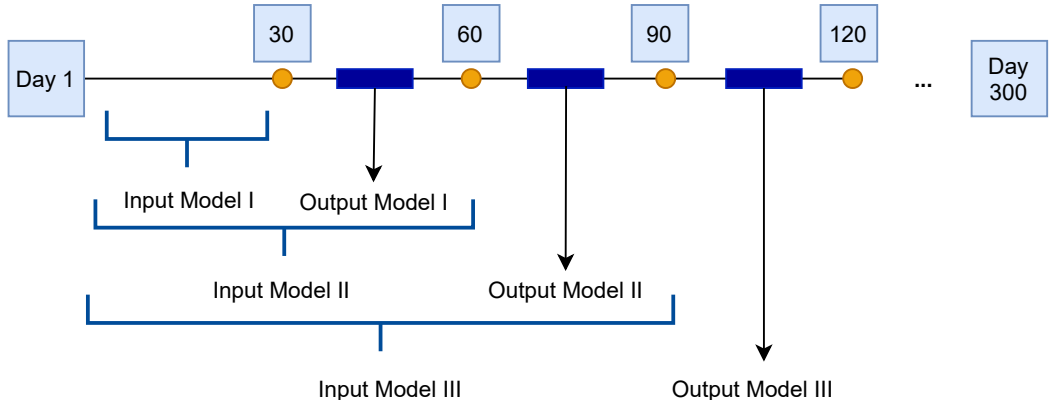}
  \caption{For evaluating the models we defined eight milestones (represented with orange circles). The performance of the methods was calculated as the average of all the evaluations.}
  \label{fig:evaluation2}
\end{figure}

This continuous training setting is not usually applied in static machine learning models due to the computational burden of training an algorithm for each new training batch. Static models need re-training strategies to keep the models up to date when dealing with non-stationary or continuous learning scenarios.
In any case, the use or optimisation of re-training strategies is outside the scope of this paper.  Still, to make a fair comparison, we used the training sets (input models) from
Figure~\ref{fig:evaluation2} as a pre-train set and used the test splits (output models) from the same figure as a test-then-train set. 
Thus, prediction errors were only evaluated during the instances that are part of the test set in experiments I and II.


To calculate the average performance of the different algorithms during the pandemic, these methods were evaluated at eight points in time, which we called milestones. Each milestone represents a date on which we predicted future cases considering only previous information to that point. Figure~\ref{fig:evaluation2} illustrates the evaluation of the applied ML methods.
Each milestone's test set covers a month interval after its respective training set. Using the subset of dates given by each monthly milestone, we created samples that contain train 
and test data for the subsequent experiments. 

\begin{figure}[h]
  \centering
  \includegraphics[width=0.85\linewidth]{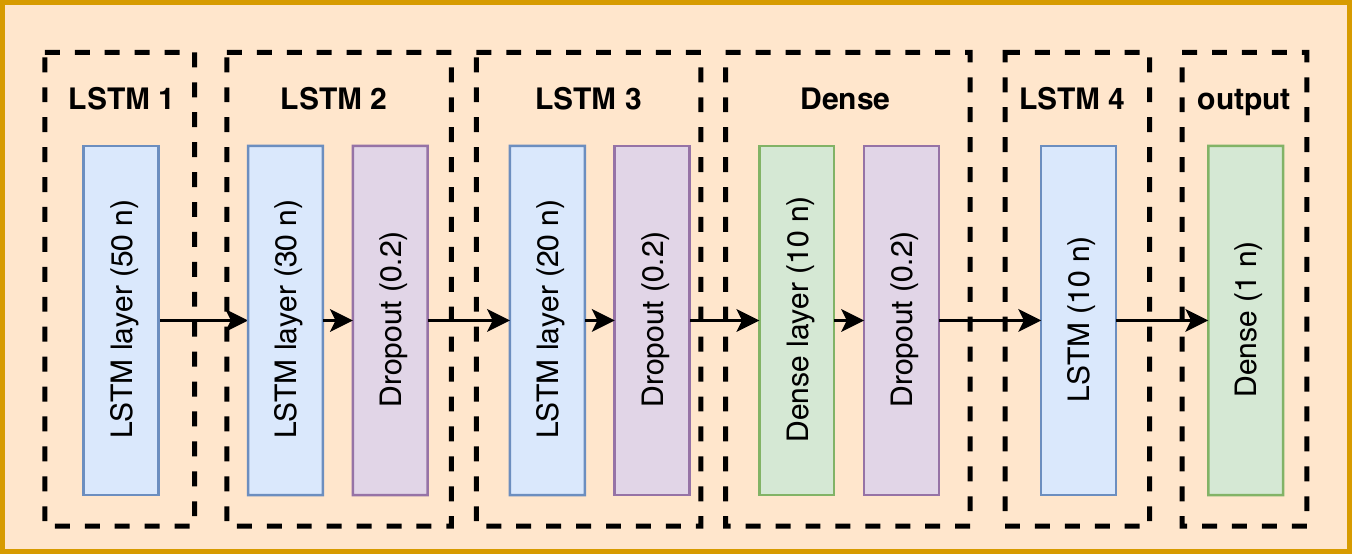}
  \caption{LSTM architecture used. The input layer receives the feature set and consist of 50 LSTM cells. The network has a total of four intermediate layers, three of them applying dropout at 0.2, and a final output dense layer.}
  \label{fig:LSTM_architecture}
\end{figure}

Since the number of models was four hundred models, eight milestones per fifty models, all algorithms are implemented using their default vanilla configuration. For the LSTM, we used the architecture from Figure \ref{fig:LSTM_architecture}. A validation set for back-propagation was created using the last ten days of the training set at each respective milestone for the LSTM only. 
The LSTM was trained for 500 epochs in both experiments. However, the batch size and patience were different in the SC, and MC approaches.

The \textit{batch size} refers to the number of training examples used in one iteration when training the LSTM sequentially. \textit{Patience} represents the number of epochs to wait before early stop training the algorithm if the model does not lower its error. 

\begin{itemize}
    \item In experiment I, 
we defined a batch size of 10 examples (one example per day) and a \textit{patience} of 20 examples for the LSTM.
    \item In experiment II, we defined a batch size of 500 examples (10 days multiplied by 50 countries) and a \textit{patience} of 1000 examples (20 days multiplied by 50 countries). 
\end{itemize}
This methodology of dividing the dataset into milestones and calculating the error as the average of the milestones was used throughout all the experiments.

The performance of the models was measured using the Root Mean Squared Error (RMSE), the Mean Absolute Error (MAE), and the Mean Absolute Percentage Error (MAPE) according to the state-of-the-art metrics for regression \cite{botchkarev2018performance}.  The results are calculated by comparing the model's predictions to the target feature in the test set (or test-then-train set in incremental learners).

The reader must note that the algorithms used may perform differently in each of these metrics. On the one hand, MAE involves the sum of the absolute values of the errors to obtain the ‘total error’. On the other hand, RMSE obtains the ‘Total square error’ as the sum of the individual squared errors; that is, each error influences the total in proportion to its square rather than its magnitude. Large errors, as a result, have a relatively more significant influence on the total square error than make the lower errors~\cite{willmott2005advantages}, which does not occur using MAE.
Finally, while MAE is the average magnitude of error produced by a model. MAPE describes how far the predictions of a model are from their corresponding outputs on average. MAPE allows comparing forecasts of different series in different scales as it is expressed in percentage-like terms. Results from MAPE can also differ with MAE as the first is undefined for data examples where the target or prediction value is zero. Thus, MAPE would be higher for an algorithm compared to other metrics if the target values are close to zero. 
This paper considers MAPE as the main evaluation metric mainly because it is a unit free metric and reports percentual-like terms.

Section~\ref{section:experiments} shows and discusses the results of three experiments performed in this work.
Experiment I is focused on models trained for a single-country (SC). Each algorithm is trained and predicts at eight different points in time, as explained in Figure~\ref{fig:evaluation2}. The MAPE, MAE and RMSE metrics are then calculated for all predictions.
Experiment II applies the same process and periods as Experiment I, but each algorithm is trained with a dataset covering all 50 countries. Each training, 
and test set includes data from all countries, as shown in Figure \ref{fig:country_selection}. Rows in these datasets are sorted first by date and then by country to respect the chronicle order of the time series for the different training, test splits and batches already mentioned.
To compare the SC results with the multi-country (MC) results from Experiment 2, the errors from the 50 models trained for a single-country are averaged. Another difference between Experiment I and Experiment II relies on the batch size for the incremental and sequential learners. Batches are time-wise for a set of days. Thus, a batch of data during training or testing is 50 times greater in Experiment II because we are training the models with data from 50 countries rather than a single one. Finally, we compare the results from the SC with the MC approach. Experiment III aims to show the benefits of using a prequential evaluation in incremental learners compared to the hold-out from experiments I and II.

\subsection{Experimentation}\label{section:experiments}
In this section, we show the results in different tables and plots for the three different experiments. We also add a subsection in which we compare the performance of the single country approach with the Multiple-Country approach. Then, we conduct the third experiment using prequential training for the incremental methods. Lastly, we perform statistical tests to compare the performance of the models.



\subsubsection{Experiment I: Single-Country training}
This experiment trains the supervised ML models with a single-country and predicts the cases for the same single-country with which the model is trained. Results are obtained averaging eight points called milestones for which predictions are made. The mean performance of the incremental and static methods for the single country (SC) approach are shown in Tables \ref{tbl:single_countries} and \ref{tbl:top_10}.

\begin{table}[ht]
\centering
\begin{tabular}{lllll}
\toprule
	Metric & MAPE & MAE  & RMSE & Time (Sec)   \\
	\midrule
	LSTM            	& 1.732   & 3053.884  & 3311.373   & \textbf{20.525} \\
	Gradient Boosting	& 1.838   & 1635.620  & 1821.918   & 0.147 \\
	Decision Tree    	& 1.862   & 1713.391  & 1976.576   & 0.005 \\
	Random Forest    	& 2.124   & 2031.514  & 2187.974   & 0.197 \\
	Bayesian Ridge   	& 7.501   & 5934.957  & 7165.991   & 0.011 \\
	HAT*       		 & 17.802  & 11502.321 & 15653.924  & 0.146 \\
	HT*                	& 18.467  & 13548.34  & 19412.316  & 0.115 \\
	ARF*            	& 28.084  & 17336.117 & 23923.305  & 2.909 \\
	PA*                	& 43.646  & 91809.182 & 111570.093 & 0.002 \\
	Linear SVR       	& 51.521  & 31370.912 & 37518.076  & 0.02 \\
\bottomrule
\end{tabular}
\caption{\label{tbl:single_countries} Models trained with the single-country (SC) approach. Results show the mean average of 50 countries. The symbol (*) indicates it is an incremental method. 
}
\end{table}

\begin{figure}[ht!]
  \centering
	\includegraphics[width=0.75\linewidth, trim=0 27 5 0, clip]{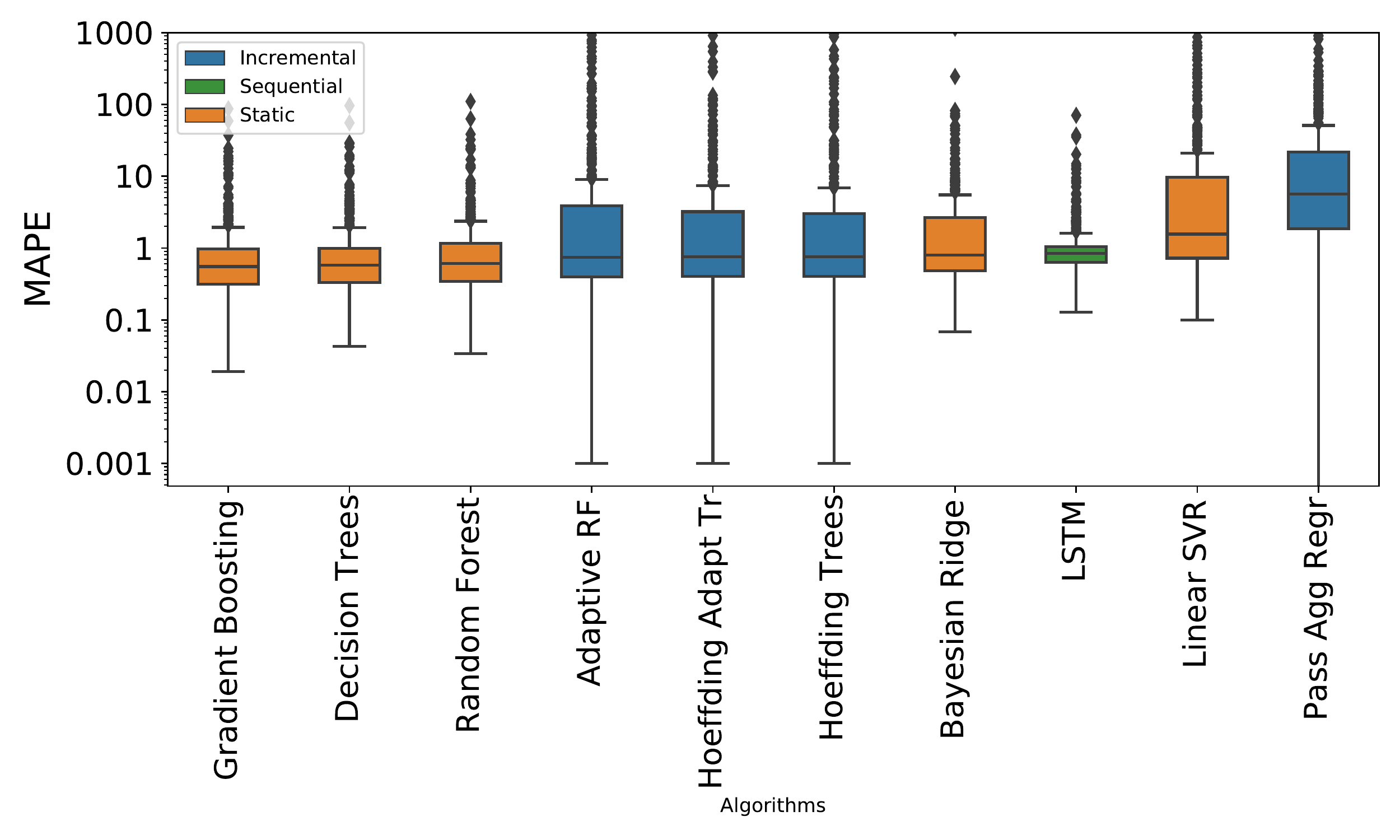}
  \caption{Boxplot for MAPE per algorithm for the single-country approach (SC). Results aggregate 50 experiments for different single countries from Table~\ref{tbl:single_countries}.}
  \label{fig:MAPE_top_countries}
\end{figure}

Table \ref{tbl:single_countries} shows the average MAE, MAPE, RMSE metric and time in seconds for the 50 countries with the most significant number of cases on the 30th of November 2020 when running over a single country. It can be seen how the LSTM algorithm outperforms the rest of the algorithms in terms of MAPE. The LSTM algorithm is followed by the static decision trees and ensembles; Gradient Boosting and Random Forest. The same results ordered by MAPE can be seen in the box-plot in Figure~\ref{fig:MAPE_top_countries}.

Table \ref{tbl:top_10} shows the average per algorithm in each of the ten countries most affected by COVID-19. The result is the average of the MAPE at eight points (milestones) between incremental and non-incremental methods. It is noticeable how the best performing algorithms in many of these countries align to the ones in Table~\ref{tbl:single_countries}.

Albeit the top performers can change in different countries, it is clear how the static methods show an overall performance higher than the incremental methods.

The majority of the algorithms in Table~\ref{tbl:single_countries} has a mean run time per milestone and country lower than 1 second for predicting each country. 
Therefore, none of these algorithms presents issues to work on a real-time setting with the current training set sizes.

\begin{figure}[ht!]
  \centering
	\includegraphics[width=0.75\linewidth, trim=0 27 5 0, clip]{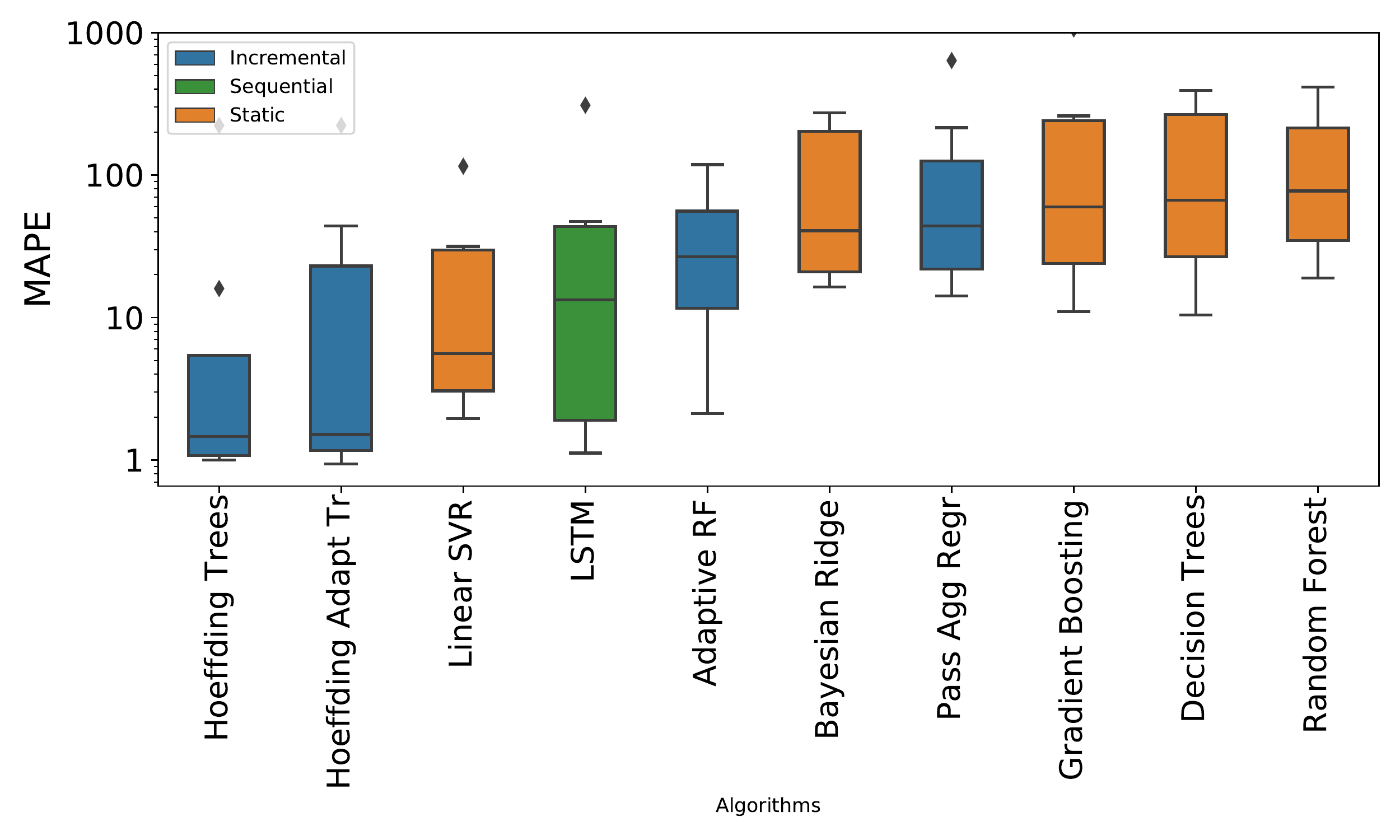}
	\caption{MAPE per algorithm 
	for the multi-country experiment (MC) covering the 50 countries from Figure~\ref{fig:MAPE_top_countries}.}
  \label{fig:MAPE_concated_countries}
\end{figure}

\begin{table*}
\centering
\begin{tabular}{lcccccccccc}
\toprule
	Country & USA & India & Brazil & Russia & France & Spain & UK & Italy & Mexico & Germany \\
\midrule
	Gradient Boosting & 0.233 & 0.283 & 0.25 & 0.256 & 0.696 & 1.066 & 0.586 & 1.37 & 0.212 & 0.656\\
	Decision Tree & 0.245 & 0.279 & 0.249 & 0.31 & 0.719 & 0.733 & 0.536 & 1.462 & 0.229 & 0.652\\
	Random Forest & 0.282 & 0.395 & 0.332 & 0.354 & 0.831 & 1.125 & 0.669 & 1.87 & 0.266 & 0.844\\
	LSTM & 0.501 & 0.717 & 0.496 & 0.628 & 0.941 & 0.95 & 0.926 & 1.809 & 0.464 & 0.994\\
	Bayesian Ridge & 0.837 & 1.647 & 1.117 & 9.784 & 1.848 & 3.91 & 1.207 & 3.894 & 2.36 & 3.034\\
	HAT* & 8.893 & 1.51 & 2.065 & 4.542 & 14.122 & 164.272 & 1.414 & 7.836 & 0.709 & 25.604\\
	HT* & 10.534 & 1.488 & 2.157 & 5.326 & 19.904 & 227.635 & 7.218 & 26.372 & 0.627 & 24.578\\
	ARF* & 15.386 & 1.063 & 3.65 & 7.37 & 26.189 & 211.315 & 6.85 & 10.055 & 2.004 & 60.621\\
	Linear SVR & 21.624 & 2.987 & 6.834 & 22.171 & 49.861 & 84.72 & 66.339 & 106.155 & 3.884 & 234.018\\
	PA* & 76.28 & 7.672 & 12.292 & 48.007 & 156.88 & 110.016 & 46.511 & 41.014 & 8.584 & 145.672\\
\bottomrule
\end{tabular}
\caption{\label{tbl:top_10}SC approach. Mean results predicting for each of the top 10 countries over the 8 milestones. 
}
\end{table*}

Finally, the values of the MAPE, MAE and RMSE are not completely correlated for all the algorithms. For instance, in Table~\ref{tbl:single_countries} Gradient Boosting and the Decision Tree are the algorithms with the lower RMSE, followed by Random Forest. RMSE penalises the degree of the deviation under or over predicting since this is based on the calculation of the total squared error ~\cite{willmott2005advantages}. Thus, Random Forest and Gradient Boosting may under-predict or over-predict to a lower degree than the LSTM model.
In any case, there can be significant variations in the scale of new cases across different countries. Thus, measures like MAE and RMSE are inappropriate because the countries with the largest number of cases could dominate the comparisons~\cite{RePEc:eee:intfor:v:4:y:1988:i:4:p:515-518}. As clarified in the experimental methodology, to handle this, in this work we consider MAPE as the main evaluation metric.

Therefore, while Gradient Boosting and the Decision Tree are the algorithms with the lowest Mean Absolute Error, we also consider that LSTM is one the best performers for single-country experiments in the context of the COVID-19 crisis since it obtains the lowest average MAPE across all milestones (see Table~\ref{tbl:single_countries}). 

\subsubsection{Experiment II: Multiple-Countries training}

This second experiment predicts over the same 50 countries at the same eight points as Experiment I, but this time we train the model with 50 countries rather than training it with a single country as in Experiment I.

The results from Table \ref{tbl:concatenated_countries} can be compared to Table \ref{tbl:single_countries}. In the MC experiment in Table~\ref{tbl:concatenated_countries}, Support Vector and the tree-based incremental models (HT, HAT and ARF) obtain the lowest MAPE across the eight-time points.  Figure~\ref{fig:MAPE_concated_countries} shows how HT and HAT have a lower median MAPE than Support Vector Regression. 
According to Figure~\ref{fig:MAPE_concated_countries}, HT and HAT seem the most reliable predictor for the MC experiment, as they offer the lowest medians, and most of their experiments fall into a normal distribution. Using the median MAPE rather than its mean as a comparative metric is helpful since the value of the mean can be distorted by the outliers.

Regarding running times (see Table~\ref{tbl:concatenated_countries}), Support Vector Regression proves to be the most cost-effective solution 
 across the eight milestones. Support Vector Regression obtains one of the three lowest MAPEs and is the fourth fastest method of all the algorithms benchmarked in experiment II.

\begin{table}[ht!]
\centering
\begin{tabular}{lcccc}
\toprule
	Metric & MAPE & MAE & RMSE & Time(sec)\\
\midrule
	Linear SVR & 24.348 & 4577.227 & 12823.977 & 1.907\\
	HT* & 30.874 & 5495.883 & 17380.819 & 7.265\\
	HAT* & 36.171 & 4802.677 & 13945.722 & 21.234\\
	ARF* & 41.651 & 4599.165 & 14150.65 & 163.506\\
	LSTM & 53.74 & 3089.347 & 8271.583 & 138.567\\
	Bayesian Ridge & 107.16 & 2943.008 & 7671.23 & 0.036\\
	PA* & 136.261 & 40040.947 & 130604.297 & 0.005\\
	Random Forest & 138.922 & 3196.719 & 7739.082 & 3.106\\
	Gradient Boosting & 215.72 & 3052.013 & 7499.656 & 6.858\\
	Decision Tree & 254.97 & 3662.684 & 8764.345 & 0.455\\
\bottomrule
\end{tabular}
\caption{\label{tbl:concatenated_countries}Average of the 50 predictions results using multiple-countries (MC) to train each model ordered by MAPE. The symbol (*) indicates it is an incremental method.}
\end{table}

Figures~\ref{fig:MAPE_concated_countries} and~\ref{fig:MAPE_top_countries} are both ordered by the median of the MAPE values in eight-time points.
While LSTM, Decision trees, Gradient Boosting and Random Forest offer overall better results in SC, the incremental learners have better performance in terms of MAPE in the MC approach. We believe that the dominance of the incremental learners in the results of the MC approach is because these can handle complex scenarios better, due to their sequential training nature. We give more details on these thoughts in subsection~\ref{section:comparison}.

\begin{figure*}[ht!]
  \centering
	\includegraphics[width=0.9\linewidth]{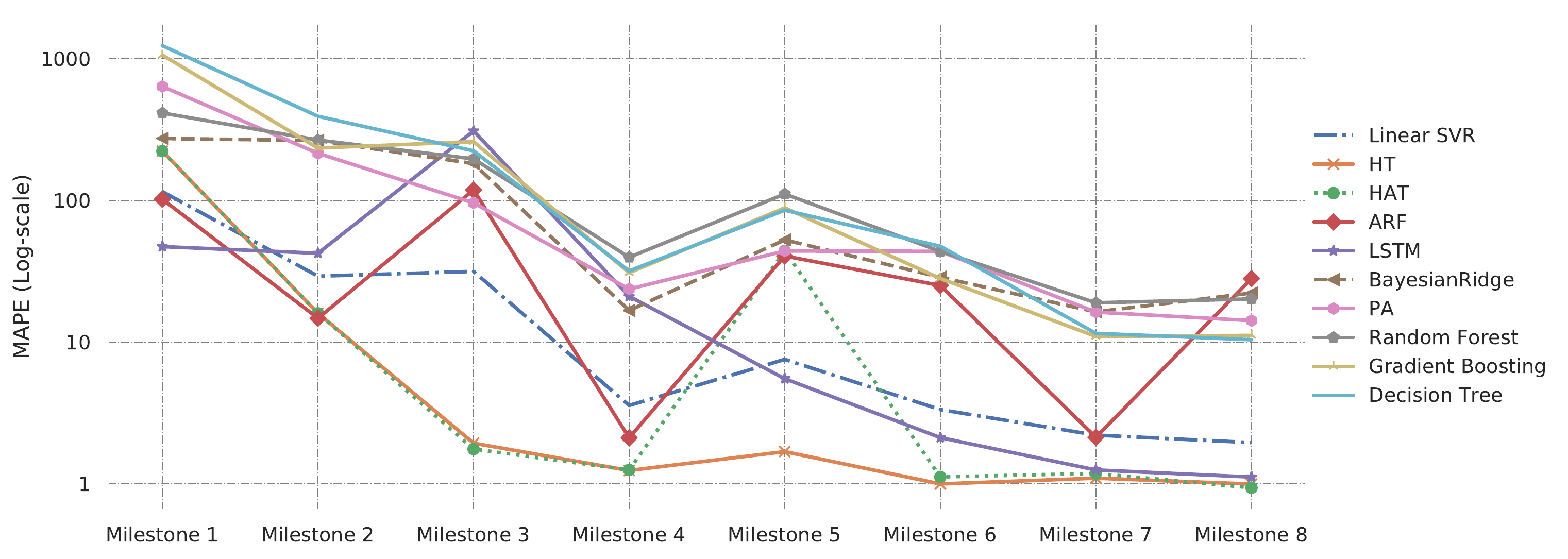}
	\caption{Evolution of MAPE per algorithm per milestone for the multi-country approach. The right-hand-side legend is sorted by the mean MAPE, as Table~\ref{tbl:concatenated_countries}.}
  \label{fig:MAPE_per_milestone}
\end{figure*}

The evolution of the MAPE obtained by each model per time-point can be seen in Figure~\ref{fig:MAPE_per_milestone}. It can be seen how the incremental learners HT and HAT are the best performers in the last milestone, followed by the LSTM, which shows a smooth evolution across the milestones. The next best performer in the last milestone is Linear SVR, which shows the lowest MAPE  mean due to its good performance for less training data in the initial milestones.

\subsubsection{Comparison between the single contry training and Multiple-countries training}
\label{section:comparison}
In the second experiment, all countries from Figure~\ref{fig:country_selection} were concatenated under a single dataset. The MC approach was proposed to provide both static and incremental algorithms with an augmented dataset that includes other countries' information and test their predictive ability with a more complex but broader set. The reader must note that this experiment was conducted to test the capacity of the models to handle a set of multiple countries where the curve of cases may not be aligned in all cases between different countries. We are aware that a batch for the incremental learners may feed multiple states of the COVID-19 curve due to the different countries used.

However, we consider this could be handled, for instance, by incremental ensembles like ARF.


\begin{table}[ht!]
\centering
\begin{tabular}{lcccc}
\toprule
	Metric          	& MAPE & MAE	& RMSE  \\
\midrule
	LSTM (SC)            	& 1.732   & 3053.884  & 3311.373   \\
	Gradient Boosting (SC)	& 1.838   & 1635.620  & 1821.918	\\
	Decision Tree (SC)   	& 1.862   & 1713.391  & 1976.576   \\
	Random Forest (SC)   	& 2.124   & 2031.514  & 2187.974   \\
	Bayesian Ridge (SC)   	& 7.501   & 5934.957  & 7165.991  \\
	HAT* (SC)       		 & 17.802  & 11502.321 & 15653.924  \\
	HT*    (SC)            	& 18.467  & 13548.34  & 19412.316  \\
	Linear SVR (MC) & 24.348 & 4577.227 & 12823.977 \\
	ARF* (SC)            	& 28.084  & 17336.117 & 23923.305 \\
	HT* (MC) & 30.874 & 5495.883 & 17380.819 \\
	HAT* (MC) & 36.171 & 4802.677 & 13945.722 \\
	ARF* (MC) & 41.651 & 4599.165 & 14150.65 \\
	PA*    (SC)          	& 43.646  & 91809.182 & 111570.093 \\
	Linear SVR (SC)       	& 51.521  & 31370.912 & 37518.076 \\
	LSTM (MC) & 53.74 & 3089.347 & 8271.583 \\
	Bayesian Ridge (MC) & 107.16 & 2943.008 & 7671.23 \\
	PA* (MC) & 136.261 & 40040.947 & 130604.297 \\
	Random Forest (MC) & 138.922 & 3196.719 & 7739.082 \\
	Gradient Boosting (MC) & 215.72 & 3052.013 & 7499.656 \\
	Decision Tree (MC) & 254.97 & 3662.684 & 8764.345 \\
\bottomrule
\end{tabular}
\caption{\label{tbl:ranking}Aggregate of results from experiments I and II, ordered by their MAPE.}
\end{table}

In general, the SC approach exhibits lower MAPE values than the MC approach. Thus, SC can be seen as the best of the two approaches. The LSTM (SC) is the best performer overall across the 50 countries used in the experiments in the eight-time points.
A rank of algorithms in both SC and MC experiments by MAPE can be seen in Table~\ref{tbl:ranking}.

We believe that the MC experiment needs a model able to map non-linearity sequentially or incrementally. This is something that we think could be handled by an incremental ensemble like Adaptive Random Forest (ARF). However, ARF was designed for purely incremental scenarios, and the purpose of its drifts detectors is to replace base regressors when the data distribution requires. Thus, the hold-out, static evaluation scheme, used in this paper, acts as a constraint to this designed behaviour.
To show the ability of ARF in this scenario, in Experiment III  
we compare MC experiment results of the incremental learners, with a hold-out evaluation scheme, to a more conventional evaluation scheme from the literature of incremental learning for data streams called prequential evaluation.



\subsubsection{Experiment III: Prequential Evaluation in the Incremental Learners}
 \label{section:hold-out_vs_preq}

The previous subsection concludes that traditional machine learning models obtain the best MAPE results in the single country approach, which shows lower MAPE than the multi-country approach. However, as mentioned at the end of the subsection, incremental machine learning algorithms are often not designed for a static evaluation. Thus, the common framework to compare them to traditional machine learning methods used in this paper may be a constraint. This subsection uses a prequential evaluation scheme over the MC experiments over the same 50 countries and eight-time points to show the performance gains that these algorithms may exhibit when evaluated appropriately.

 \begin{table}[ht!]
\centering
\begin{tabular}{lcccc}
\toprule
Metric &    MAPE &    MAE     	&  RMSE \\
\midrule
ARF (prequential)  &    10.16     & 4257.957  &    16397.503  \\
ARF (hold-out)  &	41.651 & 4599.165 & 14150.65 \\
HAT (prequential) &	27.239     & 4891.071  &    18074.166  \\
HAT* (hold-out) & 36.171 & 4802.677 & 13945.722 \\
HT (prequential)  & 27.284      & 4892.321  &    18075.028  \\
HT* (hold-out) & 30.874 & 5495.883 & 17380.819 \\
PA (prequential) & 40.864     & 22228.573 &  83660.807  \\
PA* (hold-out) & 136.261 & 40040.947 & 130604.297 \\
\bottomrule
\end{tabular}
\caption{\label{tbl:preq}Mean results of MC experiments with a hold-out vs a prequential evaluation.}
\end{table}

Table ~\ref{tbl:preq} shows that all of the incremental learners (especially ARF and PA) improve their performance in terms of MAPE by using a prequential evaluation. The MAPE obtained by ARF with a prequential evaluation would make it comparable to some of the best approaches for the SC experiment (see Table~\ref{tbl:ranking}).
Nonetheless, this subsection does not intend to compare their performance to traditional approaches as the training split differs over time when using a prequential evaluation. A continuous re-training of the conventional methods would bring a computational cost that is outside the scope in this work. In any case, to compare the two different evaluation schemes over the incremental learners, we used the hold-out training split as a pre-training split only for the prequential evaluation.
All errors from Table~\ref{tbl:preq} are computed over the Test split for hold-out, and the equivalent period, which belongs to the Test\&Train set, for the prequential evaluation.

\subsection{Statistical tests}

Due to the stochastic components of many of the algorithms used ( e.g. bagging and boosting in ensembles, initialization of weights in LSTMs), we performed statistical significance tests for the averaged results. First, we perform a normal test. If the distribution is normal, we use Welch's t-test to test for statistical significance. Conversely, if the normality test is rejected, we apply the non-parametric Mann-Whitney U test.  This last is the case in the second experiment (MC) in all cases due to the size of the input population.

In experiment I, the LSTM, Gradient Boosting, Random Forest, and the Decision Tree obtain the lowest MAPE. The LSTM obtains the lowest MAPE and is statistically significant to all other approaches at p-values of 0.05, 0.1 and 0.01 to Gradient Boosting, Random Forest, and all other algorithms, respectively. 

In experiment II, Support Vector Regression obtains the lowest MAPE, all together with HT, ARF and HAT, when trained over Multiple-Countries. The difference in MAPE is not significant between them. These five algorithms (SVR, HT, HAT, ARF and SVR) obtain the lowest MAPE, with a statistical significance of 0.05 to the rest.

Thus, given results of the statistical tests, the conclusions reached in section~\ref{section:comparison} and the medians from in Figure~\ref{fig:MAPE_top_countries} and \ref{fig:MAPE_concated_countries}, we can conclude that the LSTM has been the best performing algorithm in experiment I and overall (for a hold-out evaluation), and HT, HAT, ARF, the LSTM and SVR were the best performing algorithm in experiment II.

\section{Discussion}\label{Discussion}
\label{section:discussion}

Experiments I and II compare traditional machine learning regression algorithms to Incremental Learners for 50 countries, eight-time points and a common hold-out evaluation scheme.
In their results, we see how traditional static machine learning techniques perform better than the incremental methods in the SC experiment. Models like HT and ARF are designed for data streaming scenarios and to handle large amounts of data. We see how they beat static methods when trained for a broader set in the MC approach. These tend to adapt better over time and offer the best performance in the last milestones. Furthermore, incremental methods are oriented to online scenarios and continuous adaptation.

Experiment III compares the performance of the incremental methods using the hold-out scheme from Experiments I and II to a prequential evaluation, the scheme applied by design in their relevant literature. As seen in Experiment III, the performance of models like ARF can increase from a MAPE value of 41.65 to 10.16 by applying incremental updates during the test set (prequential evaluation). This is because ARF has internal drift detectors that create and replace decision trees when a Concept Drift is detected, affecting the performance of the regressor.

While one may think that the MC approach should give enough information to most models to improve their performance, the results show the opposite. This is probably because many of these countries have very different behaviours in the evolution of COVID-19 cases and can mislead rather than help the model predict a particular country. That is to say, at the same time point, different countries may be in states of an outbreak different from each other, such as at the start or the end of a different COVID-19 wave. The presence of different states of the disease across countries, adds extra complexity over the MC approach compared to the SC approach. 

COVID-19 is an example of a Concept Drifting environment. 
This experiment evaluates the ability of the incremental algorithms to deal with different Concepts at a time, as different countries could be on different moments of waves (or even in different waves).

In the MC experiment, Support Vector regression presents lower MAPES for the first milestones. However, incremental approaches improve their performance as the training data increases. In any case, there is no statistical significance between SVR and LSTM in the MC experiment. 
The LSTM is overall the best method across experiments. Besides its computational cost (20 times the run time of incremental learners), it offers a smooth adaptation as the data set increases (see Figure~\ref{fig:MAPE_per_milestone}).

Our results show how incremental learners can obtain the lowest error for the MC experiment. ARF, the ensemble of HT, is the incremental algorithm with the lowest MAPE when using a prequential evaluation. We believe that different trees of the adaptive ensemble may be learning about different sets of countries (due to the bagging mechanism) that may perform more or less similarly and adapt continuously to any concept changes. As for adaptive single learners or static ensembles, other algorithms can adapt to different concepts by their incremental learning or by having a set of base learners for different stationarities by using a prequential scheme. 

\section{Conclusion}\label{conclusion}

The purpose of this work is to explore the suitability of online incremental machine learning algorithms to predict the evolution of the COVID-19 virus spread accurately. It is crucial to do more research to find the best strategies and methodologies to tackle new outbreaks of other potential viruses so that their effects on public health can be addressed in a better way in the future. Online incremental machine learning models have not been exploited yet to this end, and they represent a relevant alternative due to their ability to adapt to non-stationary behaviours, which are very characteristic of epidemic curves.

In this paper, we backtested the daily information about COVID-19 cases during 2020 for 50 different countries to recreate a situation in which the countries had to face the threat of the number of cases increasing and protect the economy of the country at the same time.

In this situation (a year ago), the information about the spread of the virus and the new outbreaks was very limited. Our research is valuable because of the insights of the experiments in which we compare state-of-the-art static methods versus the performance of online incremental methods for predicting the number of new cases. 

The obtained results show that our proposed approach of using incremental methods can outperform the traditional literature static methods when training for multiple countries, especially when we follow the prequential scheme. Incremental methods adapt over time and obtain lower errors in the last periods. These algorithms also show their ability to adapt to the non-stationarities exhibited in these time series. ARF obtains a MAPE error four times lower when using a prequential evaluation instead of a hold-out scheme. Across experiments I and II, the SC experiment obtains the best results. In the SC experiment, the LSTM is the algorithm with the lowest MAPE. We should highlight that since the LSTM is designed to handle data of a sequential nature and previous results from the literature, its performance is not a surprise. The LSTM is one of the algorithms that adapt better over time across the milestones in Figure~\ref{fig:MAPE_per_milestone}. In any case, even following a hold-out scheme, the incremental methods HT and HAT obtain the lowest median MAPE in the MC experiment.  

Lastly, we proved that models trained with a single-country tend to obtain lower errors (better results) and that the error trains to diminish as the models are trained with more information. This is probably because some countries are very different from each other and can misguide the classifier. For the approach of predicting single countries, incremental methods tend to obtain higher errors (worse results) than those in other state-of-the-art techniques when compared using the static scheme of train and test hold-out splits. In any case, the proposed hold-out static scheme has proved to be a constraint by design for incremental learners.

However, a broader comparison using this scheme in traditional methods needs to be carried out to compare both approaches in a continuous setting effectively. Due to the high computational cost of re-training static algorithms from scratch, this is left for future research lines.

For future work, we would like to explore a new approach. Rather than training a classifier with all the 50 countries, we train the classifier only with those that behave similarly to the one being predicted. In other words, we will calculate first a time series similarity measure such as euclidean distance, dynamic time warping, or Symbolic Aggregate approXimation (SAX) \cite{qureshi2020valve}. Then we will use a threshold to select the countries with the lower distance to the predicted country. This would allow us to perform an MC experiment where all countries are in similar states of the curve. Thus, we expect to obtain lower prediction errors by selecting similar countries and rejecting the disparate ones.

\bibliographystyle{unsrt}  
\bibliography{biblio}

\end{document}